\documentclass[12pt, a4paper, onecolumn]{article}

\usepackage{fullpage}
\usepackage{graphicx}
\usepackage{natbib}
\usepackage[english]{babel}
\usepackage[center]{caption}
\usepackage{mathtools}
\usepackage{mathptmx}    
\usepackage{latexsym}           
\usepackage{url}
\usepackage{multirow}
\usepackage{amssymb}
\usepackage{amsthm}
\usepackage{amssymb}
\usepackage{amsmath}
\usepackage{lineno}

\long\def\symbolfootnote[#1]#2{
\begingroup
	\def\thefootnote{\fnsymbol{footnote}}\footnote[#1]{#2}
\endgroup}

\begin{document}

\title{Finding Academic Experts on a MultiSensor Approach using Shannon's Entropy}

\author{Catarina Moreira\\ \small \texttt{catarina.p.moreira@ist.utl.pt}\\
\and
Andreas Wichert\\ \small \texttt{andreas.wichert@ist.utl.pt}
\and
\\Instituto Superior T\'{e}cnico, INESC-ID\\ Av. Professor Cavaco Silva, 2744-016 Porto Salvo, Portugal\\  
\\ \small The original publication is available at: Journal of Expert Systems with Applications, Elsevier\\
 \small \text{\url{http://www.sciencedirect.com/science/article/pii/S095741741300238}}
}

\date{}

\maketitle

\begin{abstract}
Expert finding is an information retrieval task concerned with the search for the most knowledgeable people, in some topic, with basis on documents describing peoples activities. The task involves taking a user query as input and returning a list of people sorted by their level of expertise regarding the user query.  This paper introduces a novel approach for combining multiple estimators of expertise  based on a multisensor data fusion framework together with the Dempster-Shafer theory of evidence and Shannon's entropy. More specifically, we defined three sensors which detect heterogeneous information derived from the textual contents, from the graph structure of the citation patterns for the community of experts, and from profile information about the academic experts. Given the evidences collected, each sensor may define different candidates as experts and consequently do not agree in a final ranking decision. To deal with these conflicts, we applied the Dempster-Shafer theory of evidence combined with Shannon's Entropy formula to fuse this information and come up with a more accurate and reliable final ranking list. Experiments made over two datasets of academic publications from the Computer Science domain attest for the adequacy of the proposed approach over the traditional state of the art approaches.  We also made experiments against representative supervised state of the art algorithms. Results revealed that the proposed method achieved a similar performance when compared to these supervised techniques, confirming the capabilities of the proposed framework.
\end{abstract}

\symbolfootnote[0]{This work was supported by national funds through FCT - Funda\c{c}\~{a}o para a Ci\^{e}ncia e a Tecnologia, under project PEst-OE/EEI/LA0021/2011 and supported by national funds through FCT - Funda\c{c}\~{a}o para a Ci\^{e}ncia e a Tecnologia, under project PTDC/EIA-CCO/119722/2010 }

\section{Introduction} \label{sec:intro}

The search for the most knowledgeable people in some specific area, with basis on documents describing people's activities, is a challenging problem that has been receiving highly attention in the information retrieval community. Usually referred to as expert finding, the task involves taking a user query as input, with a topic of interest, and returns a list of people ordered by their level of expertise towards the query topic. Although expert search is a recent concern in the information retrieval community, there are already many research efforts addressing this specific task exploring different retrieval models. 

Many of the most effective models for expert finding are mainly based in language models frameworks. The main problem of these methods is that they can only take into account textual similarities between the query topics and documents~\citep{Balog06FormalModels, Balog09LanguageModels,serdyukov08mixtures}. More recently, there have been some works proposed in the literature which address the problem of expert finding as a combination of multiple sources of evidence. Instead of only ranking candidates through textual similarities between documents and query topics, the major concern in these approaches relies in how to combine different expertise evidences in an optimal way. Many of the proposed approaches that follow this paradigm are based on supervised machine learning techniques~\citep{Yang09bole,Macdonald11aggr} or discriminative probabilistic models~\citep{Fang}. Although these methods have the advantage of being able to combine a large pool of heterogeneous data sources in an optimal way, they are not scalable to a real world expert finding scenario for the following reasons. First,  the concept of expert itself is very ambiguous, since the expertise areas of a candidate are hard to quantify and the experience of a candidate is always varying through time. Even when different people are asked their personal opinion about experts in some topic, they often disagree. Moreover, people usually identify the most influential authors as experts, ignoring new emerging ones. 

Second, supervised machine learning techniques require manually hand-labeled training data where the top experts for some topic are identified. Since these relevance judgments are based on people's personal opinions, the system will only reflect the biases of the trainers, this way identifying more influential people than experts itself. Furthermore, it is difficult to find a sufficiently large dataset, with the respective relevance judgments, which could be representative of a real world expert finding scenario. The lack of these hand labeled data constraints the system by only enabling a small subset of query-expert pairs to be trained.

In traditional information retrieval, the combination of various ranking lists for the same set of documents is defined as rank aggregation. The techniques used to combine those ranking lists, in order to obtain a more accurate and more reliable ordering, are defined as data fusion. In the literature, these fusion techniques have been heavily used in multisensor approaches for both military and non-military applications. Sensor data fusion is defined as the usage of techniques which enable the combination of data from multiple sensors in order to achieve higher accuracies and more inferences than a single sensor. These techniques are based on several computer science domains such as artificial intelligence, pattern recognition and statistical estimation~\citep{Varshney97Sensor}. When we fuse sensor data, the levels of uncertainty arise and may affect the precision of the sensor fusion process, since incoming information provides uncertain and conflicting evidence. Many previous works of the literature addressed this issue through the usage of the Dempster-Shafer theory of evidence in order to provide a better reasoning process~\citep{Wu02,li08structural}. Other authors showed  that the success of this theory of evidence could be extended to other domains as well. In information retrieval systems, for instance, the Dempster-Shafer theory can be used to effectively quantify the relevance between documents and queries~\citep{Lalmas00, Lalmas98}. 

The Dempster-Shafer theory of evidence may be seen as a generalization of the probability theory. The development of the theory has been motivated by the observation that traditional probability theory is not able to distinguish uncertain information. In the traditional probability theory, probabilities have to be associated with individual atomic hypotheses. Only if these probabilities are known we are able to compute other probabilities of interest. In the Dempster-Shafer theory, however, it is possible to associate measures of uncertainty with sets of hypotheses, this way enabling the theory to distinguish between uncertainty and ignorance~\citep{Expert91}. 

In the domain of information theory, the Shannon's entropy has been successfully used to measure the levels of uncertainty associated to some random variable. In information retrieval, since large datasets usually contain large amounts of noise and lack from relevant information, it is straightforward that when using fusion techniques there will be an increase in conflicting information from different sources of evidence and therefore the Dempster-Shafer theory of evidence plays an important role in addressing this problem. However, the current expert finding literature has been merging different evidences without taking into account the resulting conflicting information.

In this work, we propose a novel method for the expert finding task which has a similar performance to supervised machine learning approaches, but does not require any hand-labelled training data and can be easily scalable to a real world expert finding scenario, as well as any learning to rank problem. 

We suggest a multisensor fusion approach to find academic experts, where each candidate is associated to a set of documents containing his publication's titles and abstracts. In order to extract different sources of expertise from these documents, we defined three sensors: a text similarity sensor, a profile information sensor and a citation sensor. The text sensor collects events derived from traditional information retrieval techniques, which measure term co-occurrences between the query topics and the documents associated to a candidate. The profile sensor measures the total publication record of a candidate throughout his career, under the assumption that highly prolific candidates are more likely to be considered experts. And the citation sensor uses citation graphs to capture the authority of candidates from the attention that others give to their work in the scientific community. 

Each sensor will rank the candidates according to the different evidences that they collected. Most of the times will end up disagreeing between each other by considering different candidates as experts, resulting in a conflict and in a rise of uncertainty. We apply the Dempster-Shafer theory of evidence combined with Shannon's entropy to resolve the conflict and come up with a more reliable and accurate ranking list.

The main motivation in using the Dempster-Shafer theory of evidence in this problem is given by the fact that the data fusion techniques used in the expert finding literature cannot deal with uncertainty when fusing the different sources of evidence. When the results obtained by each sensor are incompatible, a method is required to resolve this conflicting information and come up with a final decision. The Dempster-Shafer theory of evidence enables this information treatment by assigning a degree of uncertainty to each sensor. This is measured through the amount of conflicting information present in all sensors. A final decision is then made using the computed degrees of belief.

We have evaluated our expert finding system in a dataset which lacks in relevant information about academic publications from the Computer Science domain and compared it with an enriched version of the same dataset. We chose both datasets in order to verify the performance of the proposed system in different scenarios where there is poor information and a lot of noise in the data as well as in situations where the dataset is complete and full of information, this way showing that our system can be scalable to any academic dataset. The main hypothesis of this work is that the Dempster-Shafer theory of evidence can provide better results than a standard rank aggregation framework, because through conflicting information, this theory can assign a degree of belief based on uncertainty levels to each sensor and come up with a final decision. 

\subsection{Contributions}

The expert finding literature is based on two main datasets, an organizational dataset, which was made available by the Text REtrieval Conference (TREC), and an academic dataset that is the DBLP Computer Science Bibliography Dataset. Many authors have performed many experiments in both datasets. The most representative works performed in the TREC dataset belong to~\citep{Balog06FormalModels,Balog09LanguageModels} and~\citep{Macdonald08Voting,Macdonald11aggr}. For the DBLP dataset, the most representative works belong to~\citep{Yang09bole}, which has made available a dataset containing only relevance judgments for the DBLP dataset, and by~\citep{Deng08formal,deng11enhanced}. 

For this paper, we could not apply our multisensor approach to the TREC dataset, because this dataset consists of a collection of web pages and mainly textual features could be extracted. In addition, the state of the art approaches that use this dataset are only based on the textual contents between the query topics and the documents. If we applied this dataset to our multisensor approach, we would only have a single sensor detecting textual events. This would be a disadvantage since there would not be any more inferences to improve the ones made by this textual sensor. The DBLP dataset, on the other hand, contains the authors' publication records, is very rich on citation links (which enable the exploration of graph structures) and contains the publications' titles and abstracts. With this dataset, we could automatically extract different sources of evidence. For this reason, we based our experiments on the DBLP Computer Science academic dataset.

The main contributions of this paper are summarized as follows:  
\begin{enumerate}

	\item A MultiSensor Approach for Expert Finding. We offer an approach that gathers different information from data and enables the combination of different sources of evidence effectively. Contrary to machine learning methods, our approach does not leverage on hand labeled data based on personal relevance judgments. Instead, given a set of publication records, our method combines the inferences made by three different sensors, forming a more accurate and reliable ranking list. In the case of machine learning techniques, this could never happen, since the system would have to be trained using the personal opinions of individuals.  Thus, the system would only reflect the biases of the trainers. In this work, we defined three different types of sensors in order to estimate the level of expertise of an author: the textual similarity sensor, the profile information sensor and the citation sensor which detects citation patterns regarding the scientific impact of a candidate in the scientific community. Since each sensor can detect various events for each candidate, we fuse the different events using a rank aggregation approach where we explore several state of the art data fusion techniques, namely CombSUM, Borda Fuse and Condorcet Fusion (detailed in Section~\ref{sec:multisensor}). The events detected by each sensor are based on the preliminary research in~\citep{moreira11epia}.
	\\*

	\item The Formalization of a Dempster-Shafer Framework for Expert Finding.  When fusing data from different sensors, each detecting different types of evidences, it is straightforward that each sensor will give more weight to different candidates according to the information that each one of them has collected. This leads to conflicting information between the sensors. Illustrating this issue with a military application, in a presence of a plane various sensors must detect if it is friend or foe. If these sensors do not agree with each other, then we have conflicting information that has to be treated separately in order to come up with a decision if the plane is in fact friend or foe. Following the multisensor literature, we decided to address this problem through a Dempster-Shafer framework allowing one to combine evidences from different sensors and arriving at a degree of belief (represented by a belief function) that takes into account all the available evidences collected by all sensors (detailed in Section~\ref{sec:dempster-shafer}). The main advantage of using this theory is that it enables the specification of a degree of uncertainty to each sensor, instead of being forced to supply prior probabilities that add to unity, just like in traditional probability theory.
	\\*
	
	\item The Usage of Shannon's Entropy formula to help uncovering the importance of each sensor. The Dempster-Shafer theory requires that we know how certain a sensor is when detecting that a candidate is an expert. In the literature this information is usually provided by the judgments of knowledgeable people. To avoid asking people their opinion about the accuracy of each sensor, we used the Shannon's entropy formula to compute the degree of belief on each different sensor. By measuring the total entropy of each sensor, we are able to provide to the Dempster-Shafer framework belief functions based in the amount of reliable information that each sensor can detect in the presence of a candidate, instead of being dependent on other people's judgments.
	\\*
\end{enumerate}

\subsection{Outline}

The rest of this paper is organized as follows: Section~\ref{sec:related-work} presents the main concepts and related works in the literature. Section~\ref{sec:multisensor} explains the multisensor framework proposed in this paper, as well as all the events each sensor can detect and the data fusion techniques used fuse all the events perceived by each individual sensor. Section~\ref{sec:dempster-shafer} formalizes the Dempster-Shafer theory of evidence. Section~\ref{sec:shannon} details the Shannon's entropy formula developed for this work. Section~\ref{sec:validation} presents the datasets used in our experiments as well as the evaluation metrics used. Section~\ref{sec:results} presents the results obtained in the experiments and a brief discussion. Finally, Section~\ref{sec:conclusions} presents the main conclusions of this work.

\section{Related Work}\label{sec:related-work}

\begin{figure}
\begin{minipage}[b]{0.5\linewidth}
\centering
\includegraphics[width=\textwidth]{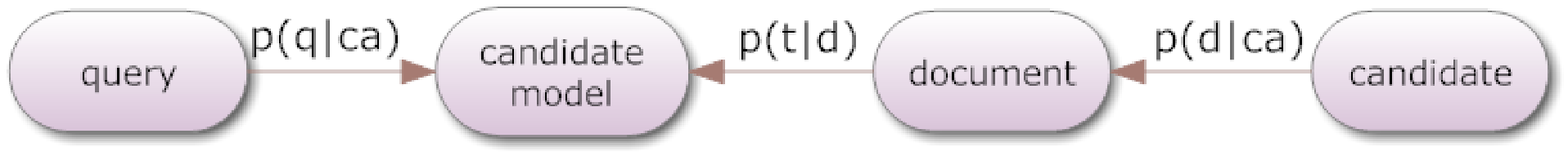}
\caption{Model 1 proposed by \citep{Balog06FormalModels} }
\label{fig:candidate-model}
\end{minipage}
\hspace{0.5cm}
\begin{minipage}[b]{0.5\linewidth}
\centering
\includegraphics[width=\textwidth]{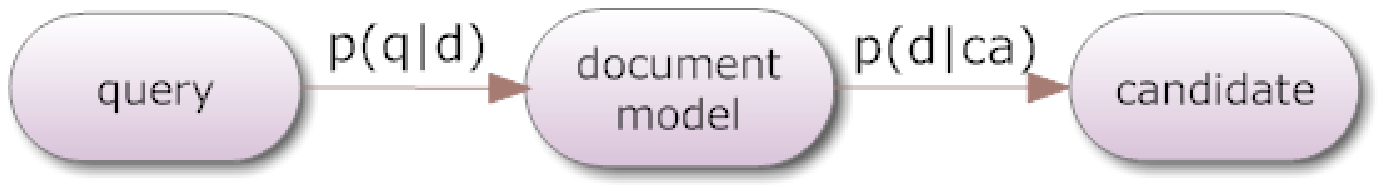}
\caption{Model 2 proposed by \citep{Balog06FormalModels} }
\label{fig:document-model}
\end{minipage}
\end{figure}

The two most popular and well formalized methods are the candidate-based and the document-based approaches. In candidate-based approaches, the system gathers all textual information about a candidate and merges it into a single document (i.e., the profile document). The profile document is then ranked by determining the probability of the candidate given the query topics, Figure~\ref{fig:candidate-model}. In the literature, the candidate-based approaches are also known as Model 1 in~\citep{Balog06FormalModels} and query independent methods in~\citep{Petkova06expertfinding}. In document-based approaches, the system gathers all documents which contain the expertise topics included in the query. Then, the system uncovers which candidates are associated with each of those documents and determines their probability scores. The final ranking of a candidate is given by adding up all individual scores of the candidate in each document~\ref{fig:document-model}. Document-based approaches are also known as Model 2 in~\citep{Balog06FormalModels} and query dependent methods in~\citep{Petkova06expertfinding}. Experimental results show that document-based approaches usually outperform candidate-based approaches~\citep{Balog06FormalModels}.

The first candidate-based approach was proposed by~\citep{Craswell01panoptic} where the ranking of a candidate was computed by text similarity measures between the query topics and the candidate's profile document.~\citep{Balog06FormalModels} formalized a general probabilistic framework for modeling the expert finding task which used language models to rank the candidates.~\citep{Petkova06expertfinding} presented a general approach for representing the knowledge of a candidate expert as a mixture of language models from associated documents. Later,~\citep{Balog09LanguageModels} and~\citep{Petkova07namedentity} have introduced the idea of dependency between candidates and query topics by including a surrounding window to weight the strength of the associations between candidates and query topics. 

In what concerns the document-based approaches, such model was first proposed by~\citep{Cao05trec} in the Text REtrieval Conference (TREC) of 2005. They proposed a two-stage language model where the first stage determines whether a document is relevant to the query topics or not and the second determines whether or not a query topic is associated with a candidate. The most well known document-based approach from the literature is Model 2, proposed by~\citep{Balog06FormalModels}. In this approach language models are used to rank the candidates according to the probability of a document model having generated the query topics. Later,~\citep{Balog09LanguageModels} explored the usage of positional information and formalized a document representation which includes a window surrounding the candidate's name. 

Methods apart from the candidate-based and the document-based approaches have also been proposed in the expert finding literature. For instance,~\citep{Macdonald08Voting} formalized a voting framework combined with data fusion techniques. Each candidate associated with documents containing the query topics received a vote and the ranking of each candidate was given by the aggregation of the votes of each document through data fusion techniques.~\citep{deng11enhanced} proposed a query sensitive AuthorRank model. They modeled a co-authorship network and measured the weight of the citations between authors with the AuthorRank algorithm~\citep{liu05networks}. Since AuthorRank is query independent, the authors added probabilistic models to refine the algorithm in order to encompass the query topics.~\citep{serdyukov08mixtures} have proposed the person-centric approach which combines the ideas of the candidate and document-based approaches. Their system starts by retrieving the documents containing the query topics and then ranks candidates by combining the probability of generation of the query by the candidate's language model.

More recently,~\citep{Macdonald11aggr}  proposed a learning to rank approach where they created a feature generator composed of three components, namely a document ranking model, a cut-off value to select the top documents according the query topics and rank aggregation methods. Using those features, the authors made experiments with the AdaRank listwise learning to rank algorithm, which outperformed all generative probabilistic methods proposed in the literature.~\citep{moreira11epia,moreira11msc} have also explored different learning to rank algorithms to find academic experts, where they defined a whole set of features based on textual similarities, on the author's profile information and based on the author's citation patterns. \citep{Fang} proposed a learning framework for expert search based on probabilistic discriminative models. They defined a standard logistic function which enabled the integration of various sets of features in a very simple model. Their features included, for instance, standard language models, document features (ex. title containing query topics), proximity features, etc.

The Dempster-Shafer theory of evidence has been widely used in the literature, specially in the sensor fusion domain. For instance,~\citep{li08structural} used a set of artificial neural networks to identify the degree of damage of a bridge. They applied the Dempster-Shafer theory together with Shannon's entropy to combine the events detected by the artificial neural networks in order to address the uncertainties arised in each network.~\citep{Wu02} formulated a general framework for context-aware (i.e., computers trying to understand our physical world). In their work, they used a set of sensors  in order to generate fragments of context information. The Dempster-Shafer theory of evidence was used to fuse the information from the various sensors and to manage the uncertainties as well as resolving the conflicting information between them. Another example of the application of this framework is in e-business with the work of~\citep{Yu05DS}. The authors proposed a modification of this framework and developed the hybrid Dempster-Shafer method to estimate the reliability of business process and quality control. Their hybrid Dempster-Shafer theory was based on entropy theory and information of co-evolutionary computation. Their results showed significant improvements over the standard Dempster-Shafer framework.

The rank aggregation framework is often used together with data fusion methods that take their inspiration on voting protocols proposed in the area of statistics and in the social sciences.~\citep{Riker88} suggested a classification to distinguish the different existing data fusion algorithms into two categories, namely the positional methods and the majoritarian algorithms. Later,~\citep{Fox94Combination} have also proposed the score aggregation methods.

The positional methods are characterized by the computation of a candidate's score based on the position that the candidate occupies in each ranking lists given by each voter. If the candidate falls in the top of the ranked list, then he receives a maximum score. If the candidate falls in the end of the list, then his score is minimum. The most representative positional algorithm is probably Borda Count~\citep{borda81} . Jean -Charles de Borda proposed this method in 1770 as being the election by order of merit method. Later, computer scientists mapped this method to combine data from different ranking lists, proving its effectiveness. 

The majoritarian algorithms are characterized by a series of pairwise comparisons between candidates. That is, the winner is the candidate which beats all other candidates by comparing their scores between each other. The most representative majoritarian algorithm is probably the Condorcet fuse method proposed by~\citep{Montague02condorcet}. However, there have been other proposals based on Markov chain models~\citep{Dwork01rankaggr} or on techniques from multicriteria decision theory~\citep{Farah07}. 

Finally, the score aggregation methods determine the highest ranked candidate by simply combining his ranking scores from all the participating systems. Examples of such methods are CombSUM, CombMNZ and CombANZ, all proposed by~\citep{Fox94Combination}.

In this article, we made experiments with representative state of the art data fusion algorithms from the positional, majoritarian and score aggregation approaches. Section~\ref{sec:multisensor} details the rank aggregation approaches.

\section{A MultiSensor Approach for Expert Finding}\label{sec:multisensor}

The multisensor approach proposed in this work contains three different sensors: a text sensor, a profile sensor and a citation sensor. 
The text sensor considers textual similarities between the contents of the documents associated with a candidate and the query topics, in order to build estimates of expertise. It is assumed that if there are textual evidences of a candidate where the query topics occur often, then it is probable that this candidate is an expert in the topic expressed by the query. The profile sensor considers the amount of publications that a candidate has made throughout his career. Highly prolific candidates are more likely to be considered experts. And finally, the citation sensor considers the impact of a candidate in the scientific community and also relies on linkage structures, such as citation graphs, to determine the candidate's knowledge. Candidates with high citations patterns are assumed to be experts.

The multisensor approach proposed in this paper consists in two different fusion processes: (i) one which will fuse all the events that each single sensor detected in a presence of a candidate and (ii) another which will fuse the information of the different sensors taking into account conflicting and uncertain data between them. The first fusion process will be addressed through a rank aggregation framework with state of the art data fusion techniques and the second one will be addressed as a multisensor fusion process using the Dempster-Shafer theory of evidence combined with Shannon's entropy.

\subsection{The Rank Aggregation Fusion Process}\label{subsec:rank}

\begin{figure}
\centering
\includegraphics[width=\columnwidth]{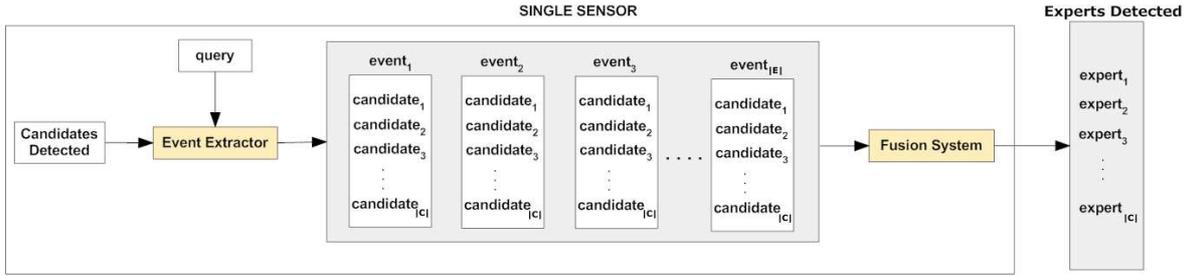}
\caption{Rank Aggregation Framework in a Single Sensor}
\label{fig:dataFusion}
\end{figure}

Rank aggregation can be defined as the problem of combining different ranking orderings over the same set of candidates, in order to achieve a more accurate and reliable ordering~\citep{Dwork01rankaggr}. Figure~\ref{fig:dataFusion} illustrates the rank aggregation framework proposed in this paper.

Given a query, the system starts by retrieving all the publication records which contain the query topics and extracts all the authors associated to those documents. These authors will be the candidates which will serve as inputs to our rank aggregation framework.

The framework is given as input a set of candidates $C=\{ c_1, c_2, ..., c_{|C|} \}$ and a query $q$ expressing a topic of expertise. Each sensor will use their own event extractor which detects a set of different events $E = \{ e_1, e_2, ..., e_{|E|} \}$ in the presence of a set of candidates. In the event extractor of every sensor, each event is responsible to order the detected set of candidates in descending order of relevance. Thus, each sensor will contain $|E|$ different ranking lists which need to be fused. A data fusion algorithm is then applied in order to combine the various ranking lists detected in each sensor. The output of this framework is a list of candidates ordered by their expertise level towards the query topic.

The events detected by each sensor are similar to the features proposed in the preliminary research work~\citep{moreira11epia} and therefore will not be detailed. Table~\ref{tab:features} summarizes all the events that can be detected by each sensor.

\begin{table}
\centering
\begin{tabular}{|l|l|l|c|c|} 
	\hline
	 {\bf Type} & {\bf Event} \\  
	\hline
	\multirow{8}{*}{{\bf Text Sensor}}	 & Query Term Frequency \\
																&	Inverse Document Frequency \\
																&	Document Length \\ 
																& Number of Unique Authors in the Documents containing query topics \\
																& Aggregated/Averaged/Maximum Okapi BM25 of documents \\
																& Aggregated/Averaged/Maximum Jaccard Coefficient of documents\\
																& Aggregated/Averaged/Maximum Okapi BM25 of conferences/journals \\
																& Aggregated/Averaged/Maximum Jaccard Coefficient of conferences/journals \\
	\hline				
	\hline
	\multirow{6}{*}{{\bf Profile Sensor}} 	& Number of Publications with(out) the query topics 	 \\ 
																		& Number of Journals with(out) the query topics 	 \\
																		& Years Since First Publication/Journal with(out) the query topics 	 \\ 
																		& Years Since Last Publication/Journal with(out) the query topics 	 \\ 
																		& Years Between First and Last Publication/Journal with(out) query topics \\
																		& Average Number of Publications/Journals per year \\
																
	\hline
	\hline
	\multirow{17}{*}{{\bf Citation Sensor}} 								& Number of Citations for papers with(out) the query topics	\\
																									& Average Number of Citations for papers containing the query topics \\
																									& Average Number of Citations per Year for papers containing the query topics\\
																									& Maximum Number of Citations for papers containing the query topics \\ 
																									& Number of Unique Collaborators in the author's publications\\
																									& Hirsch index of the author~\citep{hirsch05index}\\
																									& Hirsch Index of the author considering the query topics\\
																									& Contemporary Hirsch Index of the author~\citep{Sidiropoulos07index}\\
																									& Trend Hirsch Index of the author~\citep{Sidiropoulos07index}\\
																									& Individual Hirsch Index of the author~\citep{batista06index}\\
																									& G-Index of the author~\citep{Egghe06index}\\
																									& A-Index of the author~\citep{Jin07Aindex}\\
																									& E-Index of the author~\citep{ZhangEIndex}\\
																									& Aggregated / Average PageRank of the author's publications \\
	
	\hline
\end{tabular}
\caption{The various sets of events detect in each different sensor}
\label{tab:features}
\end{table}

Given that every single sensor will deal with the same information type (either textual, profile or citation) the conflicts when merging each event will be very low or inexistent. For instance, considering textual events, if the query topics occur very often in the candidates documents, then all the events detected by the text sensor (term frequency, inverse document frequency) will also be high. The Demspter-Shafer theory, on the other hand, will lead to a combinatorial explosion in both processing and memory requires if applied so many times. Besides, given that inside each sensor, the conflicts are very low, the Dempster-Shafer theory of evidence collapses into traditional probability theory not offering any advantages to the process.

In this article, we made experiments with representative data fusion algorithms from the information retrieval literature, namely CombSUM, Borda Fuse and Condorcet Fusion. They are described as follows.

The CombSUM score of a candidate $c$ for a given query $q$ is the sum of the normalized scores received by the candidate in each individual event, and is given by Equation~\ref{eq:CombSUM}.
\begin{equation}
CombSUM(c,q) = \sum_{e \in E} score_{e}(c, q)
\label{eq:CombSUM}
\end{equation}

The Borda Fuse positional method was originally proposed by~\citep{borda81}, in 1770, as being the election by order of merit method used in the social voting theory. It determines the highest ranked expert by assigning to each individual candidate a certain number of votes which correspond to its position in a ranked list given by each feature. Generally speaking, if a given candidate $e_j$ appears in the top of the ranking list, it is assigned to him $|E|$ votes, where $|E|$ is the number of experts in the list. If it appears in the second position of the ranked list, it is assigned $|E|-1$ votes and so on. The final borda score is given by the aggregation of each of the individual scores obtained by the candidate for each individual feature. 

The Condorcet Fusion majoritarian method was originally proposed by~\citep{Montague02condorcet} in the scope of the social voting theory. The condorcet fusion method determines the highest ranked expert by taking into account the number of times an expert wins or ties with every other candidate in a pairwise comparison. To rank the candidate experts, we use their win and loss values. If the number of wins of an expert is higher than another, then that expert wins. Otherwise, if they have the same number of wins, then we untie them by their loss scores. The candidate expert with the smaller number of loss scores wins. If the candidates have the same number of wins and losses, then they are tied~\citep{Bozkurt07fusion}.

\subsection{The MultiSensor Fusion Process}

\begin{figure}
\centering
\includegraphics[width=0.7\columnwidth]{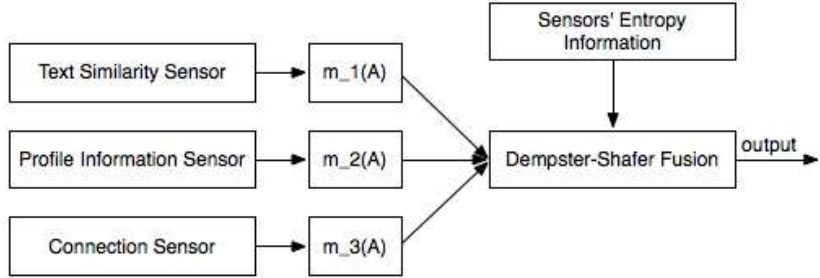}
\caption{The multisensor fusion process}
\label{fig:d-s}
\end{figure}

The multisensor fusion process is responsible to compute all the mass functions required by the Dempster-Shafer framework and to compute the amount of information that each sensor will contribute to the system through Shannon's entropy formula. This process is illustrated in Figure~\ref{fig:d-s}.

After detecting and fusing the events detected by a set of candidates, each sensor will have a ranking list where the candidates, that they believed to be experts, are in the top. However, most of the times, the top experts in these lists do not agree between each other, resulting in a conflict which needs to be treated in order to come up with a final decision. The Dempster-Shafer is then applied to compute mass functions of each sensor, $m(a)$. With this information, every sensor will have a degree of belief, enabling the fusion process. This process is detailed n Sections~\ref{sec:dempster-shafer} and~\ref{sec:shannon}.

\subsection{Example}\label{sec:ex1}

When a user submits a query with a topic of his interest, each sensor is responsible to detect specific events regarding the relation between a candidate and the query. To give an illustrative example of how the proposed system works, let us assume that a user wants to know the top experts in {\it Information Retrieval}. The first step of our system is to retrieve all the authors that have the query topics in their publication's titles or abstracts. For simplicity, let us assume that the system only found three authors with such terms: author$_1$, author$_2$ and author$_3$.

Each author has a set of documents associated with them. Each sensor is responsible to detect different types of information in those documents. The textual sensor will detect various events such as term frequency, inverse document frequency, BM25, etc. The profile sensor on the other hand, is responsible to detect the total publication record of the candidate. And the citation sensor must detect events such as the number of citations of the candidate's work, number of co-authors, etc. Each sensor can detect various events and in each event a score will be assigned to the author representing the author's knowledge towards the query topics.  Table~\ref{tab:author_sensor} shows the scores that each sensor detected in the author's documents for only two events.

\begin{table}
\resizebox{\columnwidth}{!} {
\begin{tabular}{ |l c c c c | c c c c | c c c c |}
\hline
~~ &\multicolumn{4}{ c |}{{\bf Text Sensor}} &\multicolumn{4}{ c | }{{\bf Profile Sensor}} &\multicolumn{4}{ c| }{{\bf Citation Sensor}}     \\  
\hline
Authors 		& TF	& TF$_{norm}$ & BM25	 & Bm25$_{norm}$ & Pubs & Pubs$_{norm}$ & Journ & Journ$_{norm}$ & Cits  & Cits$_{norm}$ & CitsQT & CitsQT$_{norm}$\\
author$_1$	& 9990 	& 1.0000		   & 1057 	 & 0.9440		   & 70     & 0.5769	     	   & 10 	 & 0.1200	   & 903   & 0.4929		& 266      & 0.0000 	\\
author$_2$	& 9202	& 0.2032		   & 1064 	 & 1.0000		   & 25     & 0.0000	     	   & 7       & 0.0000	   & 417   & 0.0000		& 397      & 0.5928  	\\
author$_3$	& 9001 	& 0.0000		   & 939   	 & 0.0000		   & 103   & 1.0000	     	   & 32     & 1.0000	  	   & 1403 & 1.0000		& 487      & 1.0000 	\\
  \hline
\end{tabular}
}
\caption{Scores and normalized scores for two events detected by each sensor. The normalized scores were computed using the min-max normaliation technique. $TF$ stands for term frequency, $BM25$ is a document scoring functio, $Pubs$ is the total number of publications, $Jour$ is the total number of journals, $Cits$ is the total number of citations and $CitsQT$ is the number of citations received by publications containing the query topics.}
\label{tab:author_sensor}
\end{table}

For simplicity, in this example, we will fuse the data using the CombSUM technique. In such approach, it is necessary to normalize all the scores detected by each sensor so that they range between $0$ and $1$. Then, the final score of an author is simply given by the sum of all the normalized scores that he has obtained in each event. 

\begin{table}
\centering
\begin{tabular}{ |l c | l c | l c|}
\hline
{\bf Authors} 	& {\bf Text Sensor}  &{\bf Authors} 	&{\bf Profile Sensor} 	&{\bf Authors} 	&{\bf Citation Sensor} \\  
\hline
author$_1$	& ~1.9940		 & author$_3$		&~2.0000  			&  author$_3$		& ~2.0000		 \\
author$_2$	& ~1.2032		 & author$_1$ 	&~0.6969 			&  author$_2$		& ~0.5928		\\
author$_3$	& ~0.0000 	 	 & author$_2$ 	&~0.0000    			&  author$_1$		& ~0.4929	\\
					 									
  \hline
\end{tabular}

\caption{Final scores of each sensor. Fusion using CombSUM.}
\label{tab:sensor_fusion}
\end{table}

As one can see, all three sensors disagree with each other in what concerns the final ranked list of authors. The text sensor considers $author_1$ an expert, whereas the profile and the citation sensor find $author_3$ more relevant. And in the same way, the sensors disagree between each other about the authors that hold the second and the third positions of the ranking list. If we applied CombSUM again in these three sensors, this conflicting information would not be treated. CombSUM would only sum again all the normalized scores and the output would be the final ranked list. That is, the authors which already had the highest scores would remain in the top of the list and the author with zero scores would remain with a zero score after the fusion, not giving them a chance to go up in the final ranking list. Since we are dealing with different sources of evidence with conflicting information, the next step of the algorithm is to apply the Dempster-Shafer theory of evidence together with Shannon's entropy.

\section{The Dempster-Shafer Theory of Evidence}~\label{sec:dempster-shafer}

The Dempster-Shafer theory of evidence provides a way to associate measures of uncertainty to sets of hypothesis when the individual hypothesis are imprecise or unknown~\citep{Schocken93DS}. All possible mutually exclusive hypothesis are contained in a {\it frame of discernment} $\theta$. In the scope of this work, our hypothesis will be if either an author is an expert or not. For instance, given two authors $author_1$ and $author_2$, $\theta = {author_1, author_2}$, the frame of discernment is then an enumeration of all possible combinations of these authors, that is, $2^\theta = $ \{ \{$author_1, author_2$\}, \{$author_1$\}, \{$author_2$\}, $\emptyset$ \}, performing a total of $2^\theta$ elements ($2^2=4$). The main advantage of using this theory, is that it enables the specification of a degree of  uncertainty, instead of being forced to supply prior probabilities that add to unity, just like in traditional probability theory.

The Dempster-Shafer theory of evidence enables the definition of a belief mass function which is a mapping of $2^\theta$ to the interval between 0 and 1. It can tell how relevant an author $A$ is when considering all the available evidence that supports $A$ and not any subsets of $A$. In the case of our multisensor approach, each sensor will provide a belief mass function to each author contained in the frame of discernment, by detecting the events associated with each one of them.

The belief mass function is applied to each element of the frame of discernment and has three requirements: 

\begin{itemize}

\item the mass function has to be a value between 0 and 1, $m:2^\theta \rightarrow [0,1]$

\item the mass of the empty set is zero, $m(\emptyset)=0$

\item the sum of the masses of the remaining elements is 1, $\sum_{A \in 2^\theta}{ m(A)=1 }$

\end{itemize}

When a sensor observes an author and detects the events associated with him, the probability that the observed author $A$ is relevant for some query is given by the confidence interval $[Belief(A), Plausibility(A)]$. The belief is the lower bound of the confidence interval and is defined as being the total evidence that supports the hypothesis, that is, it is the sum of all the masses of the subsets associated to set $A$.
\begin{equation}
Belief(A)= \sum_{B|B \subseteq A} m(B)
\label{eq:belief}
\end{equation}

In the same way, the plausibility corresponds to the upper bound of the confidence interval and is defined as being the sum of all the masses of the set $B$ that intersect the set of interest $A$.
\begin{equation}
Plausibility(A) = \sum_{B|B \cap A \neq \emptyset } m(B)
\label{eq:plausibility}
\end{equation}

To combine the evidences detected by two sensors $S_1$ and $S_2$, the Dempster-Shafer theory provides a combination rule which is given by Equation~\ref{eq:rule}.
\begin{equation}
m_{S_1, S_2} (\emptyset) = 0
\end{equation}
\begin{equation}
( m_{S_1} \otimes m_{S_2} )(A) = \frac{1}{1-K} \sum_{B \cap C = A \neq \emptyset}{m_{S_1}(B)m_{S_2}(C)}
\label{eq:rule}
\end{equation}

In the above formula, $K$ measures the amount of conflict between the two sensors and is given by Equation~\ref{eq:k}.
\begin{equation}
K = \sum_{B \cap C = \emptyset} m_{S_1}(B)m_{S_2}(C)
\label{eq:k}
\end{equation}

In our approach, the mass functions are used to represent the author's relevance towards the query topics, however the Dempster-Shafer theory requires that we know how certain a sensor is when  detecting that a candidate is an expert. In traditional applications of the Dempster-Shafer theory, these values are given by experts on the topic, however we did not think that asking for someone's opinion to give estimates about the accuracy of each sensor towards an author would bring a solid solution for our multisensor expert finding approach, and therefore we used a representative probabilistic formula of the information theory literature to address this issue, namely the Shannon's Entropy formula.

\section{Shannon's Entropy}\label{sec:shannon}

In information theory, entropy is defined as a measure of uncertainty of a random variable. Let $S$ be a discrete random variable representing a sensor. Assume that $E$ is the set of all events $\{e_1, e_2, ..., e_{|E|}\}$ detectable by sensor $S$ and $A$ is the set of all authors $\{ a_1, a_2, ..., a_{|A|} \}$. Let $relevantEvent(e,a)$ be a function which determines if the score detected by event $e$ for author $a$ is relevant ( that is, returns 1 if the score is bigger than zero), and $totalAuthors$, $totalEvents$ be respectively the total number of authors which are being analyzed by and the total number of events detected by the sensor. The entropy of $S$ is defined as:

\begin{equation}
H(S) = - \sum_{a \in A} \sum_{e \in E} \frac{  relevantEvent(e,a)  }{totalAuthors * TotalEvents} \log_2 \frac{  relevantEvent(e,a)  }{totalAuthors * TotalEvents} \
\label{eq:shannon}
\end{equation}

In the above formula, if $  relevantEvent(e,a) / (totalAuthors * TotalEvents)  = 1$ then the entropy is $0$, which means that the event $e$ provides consistent information for all authors and therefore there are no levels of uncertainty associated to the sensor. On the other hand, if there are high levels of uncertainty associated to the event $e$, then the maximum information associated with a sensor is given when the events are equally distributed over all authors just like described in Equation~\ref{eq:norm}.
\[ MaxH(S) = -\sum_{a \in A} \sum_{e \in E}  \frac{1}{totalAuthors * TotalEvents}\log_2\frac{1}{totalAuthors * TotalEvents} \]
\[ MaxH(S) = \log_2{(totalAuthors * TotalEvents)}~~~~~~~~~~~~~~~~~~~~~~~~~~~~~~~~~~~~~~~~~~~~~~~~~~~~~~~~~~~\]
\begin{equation}
\label{eq:norm}
\end{equation}

In conclusion, returning to the Dempster-Shafer theory of evidence, the mass function of an author $A$ detected by some sensor $S$ will be given by Equation~\ref{eq:d-s-final}, where $Fusion(A)$ represents the rank aggregation score of candidate $A$ using a data fusion algorithm. For more details in how to use the following equation, please refer to the book of~\citep{Expert91}. Section~\ref{sec:ex1} shows a detailed example in how this formula is applied in our system.

\[
  m_S(A) = \left\{ 
  \begin{array}{l l}
     \text{Fusion($A$)}			& \quad \text{if $A$ = $\{A\}$}\\
     \frac{H(S)}{maxH(S)}			& \quad \text{if $A$ = $\theta$}\\
    0		 					& \quad \text{otherwise}\\
   \end{array} \right.
 \]
\begin{equation}
\label{eq:d-s-final}
\end{equation}

\subsection{Example}\label{sec:ex2}

Continuing the example started in Section~\ref{sec:ex1}, after fusing the data in each sensor, we ended up noticing that the sensors did not agree with each other about the top experts, resulting in conflicting information that needs to be treated.

Before applying the Dempster-Shafer theory of evidence, we need to determine the importance of each sensor. If the profile sensor isn't very reliable, then when fusing data, one should take that into consideration by decreasing the scores of the authors detected in that sensor. To compute the relevance weight of each sensor, we made use of Shannon's Entropy formula described in Equation~\ref{eq:norm}.

In the previous example, in Table~\ref{tab:author_sensor} there are no zero entries in the unnormalized scores of each sensor. That means that, in this example, all sensors are equally important and therefore the Maximum Entropy for $3$ authors with $2$ non-zero detected events for each one of them is given by the following formula: 

\[MaxH(TextSensor) = \log_2{(totalAuthors*TotalEvents(TextSensor)} = \log_2 (3 \times 2) = 2.5850\]

\[MaxH(CitationSensor) = MaxH(ProfileSensor) = MaxH(TextSensor) = 2.5850\]

The Shannon's entropy formula of each sensor is given by Equation~\ref{eq:shannon} and for this example is computed in the following way:

\[H(TextSensor) = H(ProfileSensor)  = H(CitationSensor) = -  3*\frac{2}{6}\log_2 \frac{2}{6}  = 1.5850\]

At this point, we can compute the overall mass function of the sensors which is given by:

\[m(Sensor) = \frac{ H(Sensor) }{ MaxH(Sensor) } = \frac{2.5850}{1.5850} = 0.6132\]

And since the mass function requires that the sum of its elements is $1$, we need to normalize these values.

\[m(TextSensor) = m(ProfileSensor) = m(CitationSensor) = \frac{  0.6132 }{ 1.8396 } = \frac{1}{3} \]

Now, we need to add the previous CombSUM fusion results to the mass function as well, such that the sum of all elements is one. Table~\ref{tab:sensor-mass} shows such values.

\begin{table}
\resizebox{\columnwidth}{!} {
\begin{tabular}{ |l c c| l c c| l c c|}
\hline
{\bf Authors} 	& {\bf Text Sensor}	& {\bf Norm Scores}	&{\bf Authors}	&{\bf Profile Sensor}	& {\bf Norm Scores}	&{\bf Authors}	&{\bf Citation Sensor}	& {\bf Norm Scores}\\  
\hline
author$_1$	& ~1.9940 		& ~0.4118		 	& author$_3$	&~2.0000  			& ~0.4944			&  author$_3$	& ~2.0000	& ~0.4321\\
author$_2$	& ~1.2032 		& ~0.2549			& author$_1$ 	&~0.6969  			& ~0.1723 			&  author$_2$	& ~0.5928	& ~0.1281\\
author$_3$	& ~0.0000		& ~0.0000			& author$_2$ 	&~0.0000  			& ~0.0000 			&  author$_1$	& ~0.4929	& ~0.1065\\	
entropy		& ~0.6132		& ~0.3333			&entropy 	&~0.6132  			& ~0.3333 			&  entropy 	& ~0.6132	& ~0.3333\\
\hline
{\bf sum}			&				& 1					& {\bf sum}		& 					& 1					& {\bf sum}		& 			& 1 \\
\hline	
\end{tabular}
}
\caption{Final scores of each sensor. Fusion using CombSUM.}
\label{tab:sensor-mass}
\end{table}

At this point, we are ready to apply the Dempster-Shafer theory of evidence framework. We will start fusing the Text Sensor with the Profile Sensor. The mass functions of the Text Sensor are discriminated in Table~\ref{tab:m_ts} and the mass functions of the Profile Sensor in Table~\ref{tab:m_ps}. Note that these tables are in accordance with Equation~\ref{eq:d-s-final}. If $A$ is a single author, we apply the normalized scores from the rank aggregation fusion process and if $A$ is a set of authors detected by each sensor, then we apply the normalized entropy score of the sensor.

\begin{table}
\resizebox{\columnwidth}{!} {
\begin{tabular}{c c c c}
\multicolumn{4}{ c }{{\bf Text Sensor}}    \\

m(\{author$_1$\}) = 0.4118 & m(\{author$_2$\}) = 0.2549 & m(\{author$_3$\}) = 0.0000 & m(\{author$_1$, author$_2$, author$_3$\}) =  0.3333\\

\end{tabular}
}
\caption{Text Sensor mass functions}
\label{tab:m_ts}
\end{table}

\begin{table}[ht]
\resizebox{\columnwidth}{!} {
\begin{tabular}{c c c c}
\multicolumn{4}{ c }{{\bf Profile Sensor}}    \\

m(\{author$_1$\}) = 0.1723 & m(\{author$_2$\}) = 0.0000 & m(\{author$_3$\}) = 0.4944 & m({author$_3$, author$_1$, author$_2$}) =  0.3333\\

\end{tabular}
}
\caption{Profile Sensor mass functions}
\label{tab:m_ps}
\end{table}

The fusion process under the Dempster-Shafer theory of evidence framework is given through the computation of a $tableau$ given by Table \ref{tab:fusion1}. In this tableau, we perform the intersection between each element of the Text Sensor, with each element of the profile Sensor. For instance, m$_{textSensor}$(\{author$_1$\}) $\cap$ m$_{profileSensor}$(\{author$_1$, author$_2$, author$_3$\}) = \{author$_1$\} with probability $0.4118 \times 0.3333 = 0.1373$. Note that we multiply the probabilities, because the authors are considered independent between the two different sensors. The choice of an author as expert in the text sensor does not affect the choice of another author in the profile sensor. 

\begin{table}
\resizebox{\columnwidth}{!} {
\begin{tabular}{ l | c c c c c}
~~						& \{author$_1$\}(0.1723)	& \{author$_2$\}(0.0000)	&\{author$_3$\}(0.4944)		&\{author$_3$, author$_1$, author$_2$\} (0.3333)\\  
\hline
\{author$_1$\}(0.4118)		& \{author$_1$\}(0.0710)	& $\emptyset$ (0.0000)		& $\emptyset$ (0.2036) 	& \{author$_1$ \} (0.1373)	\\
\{author$_2$\}(0.2549)		& $\emptyset$ (0.0439)		& \{author$_2$\}(0.0000)	& $\emptyset$ (0.1260) 	& \{author$_2$ \} (0.0850)	\\
\{author$_3$\}(0.0000)		& $\emptyset$ (0.0000)		& $\emptyset$ (0.0000)		& \{author$_3$\}(0.0000) 	& \{author$_3$ \} (0.0000)	\\
\{author$_1$, author$_2$, 
	author$_3$\} (0.3333)	& \{author$_1$\} (0.0574)	& \{author$_2$\} (0.0000)	& \{ author$_3$ \} (0.1648) 	& \{author$_1$, author$_2$,author$_3$\} (0.1111)\\
\hline
\end{tabular}
}
\caption{Final scores of each sensor. Fusion using CombSUM.}
\label{tab:fusion1}
\end{table}

Whenever the intersection gives an empty set, then this means that there is a conflict between the sensors and the combination rule in Equation~\ref{eq:rule} must be applied. This rule is given by summing all the probabilities of all the events which ended up as an empty set and subtract it by $1$. Then, we divide each of the mass functions obtained in the tableau by this value. The following calculations demonstrate the computations of the fused mass functions using the combination rule.

\[k = 1 - (0.2036 + 0.0439 + 0.1260) = 1 - 0.3735 = 0.6265\]\\
\[m_{textSensor} \oplus m_{profileSensor}(\{ author_1 \}) = (0.0710 + 0.1373 + 0.0574) / k = 0.4241\]
\[m_{textSensor} \oplus m_{profileSensor}(\{ author_2 \}) = (0.0000 + 0.0850 + 0.0000) / k = 0.1357\]
\[m_{textSensor} \oplus m_{profileSensor}(\{ author_3 \}) = (0.0000 + 0.0000 + 0.1648) / k = 0.2630\]
\[m_{textSensor} \oplus m_{profileSensor}(\{ author_1, author_2, author_3 \}) = 0.1111/ k = 0.1772\]

What is interesting to notice in the above calculations is that no author ended up with a zero probability, although each sensor detected that some authors were irrelevant. If we did not use Shannon's Entropy to weight the importance of each sensor, author$_3$ and author$_2$ would end up with a probability of zero, meaning that these authors are completely irrelevant for the query. Shannon's Entropy enabled to give some belief in these authors, given the importance of their respective sensor, and enabled a more consistent and reliable ranking for each one of them.

Next, the same process is used to fuse the computed results with the Citation Sensor.

\begin{table}
\resizebox{\columnwidth}{!} {
\begin{tabular}{c c c c}
\multicolumn{4}{ c }{{\bf Text Sensor $\oplus$ ProfieSensor}}    \\

m(\{author$_1$\}) = 0.4241 & m(\{author$_2$\}) = 0.1357 & m(\{author$_3$\}) = 0.2630 & m({author$_3$, author$_2$, author$_1$}) =  0.1772\\

\end{tabular}
}
\caption{Text Sensor $\oplus$ Profile Sensor mass functions}
\label{tab:tex_prof_sensors}
\end{table}

\begin{table}[ht]
\resizebox{\columnwidth}{!} {
\begin{tabular}{c c c c}
\multicolumn{4}{ c }{{\bf Citation Sensor}}    \\

m(\{author$_1$\}) = 0.1065 & m(\{author$_2$\}) = 0.1281 & m(\{author$_3$\}) = 0.4321 & m({author$_3$, author$_2$, author$_1$}) =  0.3333\\

\end{tabular}
}
\caption{Citation Sensor mass functions}
\label{tab:conn_sensor}
\end{table}

\begin{table}[ht]
\resizebox{\columnwidth}{!} {
\begin{tabular}{ l | c c c c c}
~~						& \{author$_1$\}(0.1065)	& \{author$_2$\}(0.1281)	&\{author$_3$\}(0.4321)		&\{author$_3$, author$_1$, author$_2$\} (0.3333)\\  
\hline
\{author$_1$\}(0.4241)		& \{author$_1$\}(0.0452)	& $\emptyset$ (0.0543)		& $\emptyset$ (0.1833) 	& \{author$_1$ \} (0.1414)	\\
\{author$_2$\}(0.1357)		& $\emptyset$ (0.0145)		& \{author$_2$\}(0.0174)	& $\emptyset$ (0.0586) 	& \{author$_2$ \} (0.0452)	\\
\{author$_3$\}(0.2630)		& $\emptyset$ (0.0280)		& $\emptyset$ (0.0337)		& \{author$_3$\}(0.1136) 	& \{author$_3$ \} (0.0877)	\\
\{author$_1$, author$_2$, 
	author$_3$\} (0.1772)	& \{author$_1$\} (0.0189)	& \{author$_2$\} (0.0227)	& \{ author$_3$ \} (0.0766) 	& \{author$_1$, author$_2$,author$_3$\} (0.0591)\\
\hline
\end{tabular}
}
\caption{Tableau for the final combination between the TextSensor $\oplus$ ProfileSensor with the Citation Sensor.}
\label{tab:tableau_final}
\end{table}

\[k = 1 - (0.0543 + 0.1833 + 0.0145 + 0.0586 + 0.0280 + 0.0337) = 1 - 0.3724 = 0.6276\]
\[m_{textSensor} \oplus m_{profileSensor}\oplus m_{citationSensor}(\{ author_1 \}) = (0.0452 + 0.1414 + 0.0189) / k = 0.3274\]
\[m_{textSensor} \oplus m_{profileSensor}\oplus m_{citationSensor}(\{ author_2 \}) = (0.0174 + 0.0452 + 0.0227) / k = 0.1359\]
\[m_{textSensor} \oplus m_{profileSensor}\oplus m_{citationSensor}(\{ author_3 \}) = (0.1136 + 0.0877 + 0.0766) / k = 0.4428\]
\[m_{textSensor} \oplus m_{profileSensor}\oplus m_{citationSensor}(\{ author_1, author_2, author_3 \}) = 0.0591/ k = 0.0942\]

The algorithm ends by retrieving the final ranking list: author$_3$(0.4428) $>$ author$_1$(0.3274) $>$ author$_2$(0.1359)

\section{Experimental Setup}\label{sec:validation}

The multisensor approach required a large dataset containing not only textual evidences of the candidates knowledge, but also citation links. In this work, we made experiments with two different versions of the Computer Science Bibliography dataset, also known as DBLP\footnote{\url{http://www.informatik.uni-trier.de/~ley/db/}} .

The DBLP dataset has been widely used in the expert finding literature through the works of~\citep{Deng08formal,deng11enhanced} and~\citep{Yang09bole}. It has also been extensively used in citation analysis in the works of~\citep{Sidiropoulos05citation,Sidiropoulos06generalized}. It is a large dataset covering publications in both journals and conferences and is very rich in citation links.

The two versions of the DBLP dataset tested in this work correspond to the Proximity\footnote{\url{http://kdl.cs.umass.edu/data/dblp/dblp-info.html}} version and an enriched DBLP\footnote{\url{http://www.arnetminer.org/citation}} version. Proximity contains information about academic publications until April 2006. It is a quite large dataset containing more than 100 000 citation links and 400 000 authors, however it does not provide any additional textual information about the papers besides the publication's titles. On the other hand, the enriched version of the DBLP dataset, which has been made available by the Arnetminer project, is a large dataset covering more than one million authors and more than two million citation links. It also contains the publication's abstracts of more than 500 000 publications. Table~\ref{t1} provides a statistical characterization of both datasets. We made experiments with both datasets to verify the scalability of our method in the presence of datasets containing a lot of information and datasets full of noise and lacking on relevant information.

\begin{table}
\begin{center}
\resizebox{\columnwidth}{!} {
\begin{tabular}{l c c}
  
  Object																															& Proximity 										& DBLP								\\
  \hline
  Total Authors 		& ~~~~456 704~~~~						& ~~~~1 033 050~~~~		\\
  Total Publications 	& ~~~~743 349~~~~ 						& ~~~~1 632 440~~~~ 		\\
  Total Publications containing Abstract	& ~~~~0 ~~~~			& ~~~~653 514~~~~ 			\\
  Total Papers Published in Conferences~~~~~~~~~~~~~~~~~~~~~~~~~~~~~~~~~~~~~~~~~~~		& ~~~~335 480~~~~						& ~~~~606 953~~~~			\\
  Total Papers Published in Journals 																				& ~~~~270 457~~~~ 						& ~~~~436 065~~~~ 			\\
  Total Number of Citations Links 																					& ~~~~112 303~~~~ 						& ~~~~2 327 450~~~~ 		\\
  \hline
\end{tabular}
}
\end{center}
\scriptsize
\caption{Statistical characterization of the Proximity dataset and the enriched version of the DBLP dataset used in our experiments}
\label{t1}
\end{table}

To validate the different experiments performed in this work, we required a set of queries with the corresponding author relevance judgements. We used the relevant judgements provided by Arnetminer,\footnote{\url{http://arnetminer.org/lab-datasets/expertfinding/}} which have already been used in other expert finding experiments~\citep{Yang09bole,deng11enhanced}. The Arnetminer dataset comprises a set of 13 query topics from the Computer Science domain, and it was mainly built by collecting people from the program committees of important conferences related to the query topics. Table~\ref{judgements} shows the distribution for the number of experts associated to each topic, as provided by Arnetminer.
\begin{table}
\begin{center}
\resizebox{\columnwidth}{!} {
\begin{tabular}{l c l c}

~~{\bf Query Topics}								& ~~{\bf Rel. Authors}		&~~{\bf Query Topics}  ~~~~~~~~~~~~~~~						&~~{\bf Rel. Authors}	\\
\hline

 ~~Boosting (B)											& ~~46								&	~~Natural Language (NL)					& ~~41							\\
 ~~Computer Vision (CV)							&	~~176							&	~~Neural Networks	 (NN)					& ~~103						\\
 ~~Cryptography (C)									&	~~148							& ~~Ontology				 (O)					& ~~47							\\
 ~~Data Mining (DM)								&	~~318							& ~~Planning				 (P)					& ~~23							\\
 ~~Information Extraction (IE)					&	~~20							& ~~Semantic Web		 (SW)					& ~~326						\\
 ~~Intelligent Agents	(IA)						&	~~30							& ~~Support Vector Machines (SVM)	& ~~85							\\
 ~~Machine Learning	(ML)						& ~~34								& 															&									\\
\hline
\end{tabular}
}
\end{center}
\caption{Characterization of the Arnetminer dataset of Computer Science experts.}
\label{judgements}
\end{table}

Since the Arnetminer dataset contains only relevant judgements for all query topics, we complemented this dataset by adding non relevant authors for each of the query topics. Our validation set included all relevant authors plus a set of non relevant authors until we end up with a set of 400 authors. These non relevant authors were added by searching the database with the keywords associated to each topic and that were highly ranked according to the BM25 metric. Thus, the validation sets\footnote{The validation sets built for this work can be made available if requested to the authors, so other researchers can compare their approaches to ours} built for each dataset contained exactly the same relevant authors, but had different non relevant ones.

The performance of our multisensor approach was validated through the usage of the Mean Average Precision (MAP) metric and Precision at rank $k$ (P@k).
Precision at rank $k$ is used when a user wishes only to look at the first $k$ retrieved domain experts. The precision is calculated at that rank position through Equation~\ref{eq:PrecisionRank}.
\begin{equation}
P@k=\frac{r\left(k\right)}{k}
\label{eq:PrecisionRank}
\end{equation}
In the formula, $r(k)$ is the number of relevant authors retrieved in the top {\it k} positions. $P@k$ only considers the top-ranking experts as relevant and computes the fraction of such experts in the top-$k$ elements of the ranked list.

The Mean of the Average Precision over test queries is defined as the mean over the precision scores for all retrieved relevant experts. For each query $q$, the Average Precision (AP) is given by:
\begin{equation}
AP[q] = \frac{ \sum_{rn=1}^{N} (P(rn) \times rel(rn)) }{R}
\end{equation}

In the formula, $N$ is the number of candidates retrieved, $rn$ is the rank number, $rel(rn)$ returns either 1 or 0 depending on the relevance of the candidate at $rn$. $P(rn)$ is the precision measured at rank $rn$ and $R$ is the total number of relevant candidates for a particular query $q$. 

We also performed statistical significance tests over the results using an implementation of the two sided randomization test~\citep{Smucker07StatSig} made available by Mark D. Smucker\footnote{\url{http://www.mansci.uwaterloo.ca/~msmucker/software/paired-randomization-test-v2.pl}}

\section{Experimental Results}\label{sec:results}

This section presents the results of the experiments performed in this work, more specifically: 

\begin{enumerate}

\item In Section~\ref{subsec:1}, we compared our multisensor approach against a general rank aggregation framework, using only the Proximity dataset. The results of this experiment showed that the Dempster-Shafer theory of evidence combined with Shannon's entropy enables much better results than the standard rank aggregation approach.

\item In Section~\ref{subsec:2}, we determined the impact of each sensor of our multisensor approach using the Proximity dataset. Experiments revealed that the combination of the text similarity sensor together with the citation sensor achieved the best results in this dataset. Results also unveiled that  combining estimators based on the author's publication record and on their citations patterns in the scientific community achieved the poorest results.

\item In Section~\ref{subsec:3}, we repeated the experiments of our multisensor approach on an enriched version of the DBLP dataset. We demonstrated that the Dempster-Shafer theory of evidence also provides better results when used in datasets which do not have high levels of conflicting information. These results prove how general our multisensor approach can be, since it provides better results than the standard rank aggregation approach, whether using datasets with poor information (with high levels of conflict and uncertainty) or with enriched datasets (low levels of uncertainty and high levels of confidence).

\item In Section~\ref{subsec:5}, we compared our multisensor approach against representative state of the art works. Results showed that our approach achieved a MAP of more than 66\% when compared to non-machine learning works of the state of the art, becoming one of the top contributions in the literature.

\item More recently, approaches based on supervised machine learning techniques have been proposed in the literature of expert finding~\citep{Yang09bole,Macdonald11aggr}. In Section~\ref{subsec:6} we compared our multisensor approach against two supervised learning to rank techniques from the state of the art. The results obtained showed that the usage of a supervised approach does not bring significant advantages to the system when compared to our multisensor data fusion approach, concluding that this approach provides very competitive results without the need of hand-labelled data with personal relevance judgements.

\end{enumerate}

\subsection{Comparison of the MultiSesnor approach Against a General Rank Aggregation Approach using the Proximity Dataset}~\label{subsec:1}

The main hypothesis motivating this experiment is to verify if our multisensor approach, combined with the Dempster-Shafer theory of evidence, achieves better results than a standard rank aggregation approach. To validate this hypothesis, we experimented our multisensor approach with different data fusion techniques and compared it with the rank aggregation framework using the same fusion algorithms. Table~\ref{tab:proximity-d-s-results} presents the obtained results for the three proposed sensors, more specifically the text similarity sensor, the profile information sensor and the citation sensor. 
\begin{table}
\resizebox{\columnwidth}{!} {
\begin{tabular}{ l c c c c c c }
			 																			&~~~~{\bf P@5}~~~~ 		& ~~~~{\bf P@10}~~~~ 	& ~~~~{\bf P@15}~~~~ 	& ~~~~{\bf P@20}~~~~ 	& ~~~~{\bf MAP}~~~~ 	\\
\hline
{\bf MultiSensor Approach using}											&									&										&										&										&									\\
~~Dempster-Shafer + CombSUM (D-S + CSUM)							& {\bf 0.7538}		& {\bf 0.7000}			& {\bf 0.6256}			& {\bf 0.5769} 			& 0.4402					\\
~~Dempster-Shafer + Borda Fuse (D-S + BFuse)					& 0.2154					& 0.2154  					& 0.2205 						& 0.2346 						& 0.2533      		\\
~~Dempster-Shafer + Condorcet Fusion (D-S + CFusion)	& {\bf 0.7538}		& 0.6385			  		& 0.5846						& 0.5615			 			& {\bf 0.4905}  	\\			\hline
{\bf Standard Rank Aggregation Approach using} 				& 								& 									& 									& 									&			 		\\
~~CombSUM																							& 0.3385					& 0.3308						& 0.3385						& 0.3115 						& 0.3027			\\
~~Borda Fuse																					& 0.4462					& 0.4308  					& 0.4205 						& 0.3962 						& 0.3402  	    \\
~~Condorcet Fusion																		& 0.4615					& 0.4538  					& 0.3846 						& 0.3538 						& 0.2874  		\\
\hline
{\bf Relative Improvements}  													& 								& 									& 									& 									&			 			\\
~~ D-S + CFusion vs CombSUM 													&	+122.69\%*			&	+93.02\%*					&	+72.70\%					&	+80.26\%*	 				& 	+62.04\%		\\
~~ D-S + CFusion vs Borda Fuse 												&	+68.09\%*				&	+48.21\%					&	+39.02\%					&	+41.72\%					&	+44.18\%*		 \\
~~ D-S + CFusion vs Condorcet Fusion  								&	+63.34\%*				&	+40.70\%					&	+52.00\%					&	+58.71\%					&	+70.67\%*		  \\
\hline
\end{tabular}
}
\caption{Results of sensor fusion using the Dempster-Shafer framework and sensor fusion using only representative data fusion algorithms. The experiments were performed using the three proposed sensors (text, profile and citation). * indicates that the improvement of the best approach presented is statistically significant for a confidence interval of 90 \% }
\label{tab:proximity-d-s-results} 
\end{table}

The results on Table~\ref{tab:proximity-d-s-results} show that our multisensor approach using the Dempster-Shafer framework outperformed the general rank aggregation approach. When two sensors do not agree between each other, it is difficult to get a final decision whether a candidate is an expert or not. In these situations, a standard rank aggregation approach simply ignores the conflict and applies a data fusion technique to merge the scores of that candidate in both sensors.  The Dempster-Shafer theory of evidence, on the other hand, assigns a degree of uncertainty to each sensor which is measured through the amount of conflicting information present in both of them. A final decision is then made using the computed degrees of belief. This experiment shows that, when merging different sources of evidence, conflicting information should not be ignored. Thus, the Dempster-Shafer theory of evidence plays an important role in solving these conflicts and providing a final decision. In conclusion, these results support the main hypothesis of this work which so far has been ignored in the expert finding literature: when merging multiple sources of evidence, it is necessary to apply methods to solve conflicting information, this way enabling a more accurate and more reliable reasoning. The best performing data fusion technique is the majoritarian Condorcet Fusion algorithm. In our multisensor approach, Condorcet Fusion achieved an improvement of more than 70\%, in terms of MAP, when compared to the same algorithm in the standard rank aggregation approach.

\subsection{Determining the Impact of the Different Sensors in the MultiSensor Approach}~\label{subsec:2}

The data reported in the previous experiment showed that the proposed multisensor approach, combined with the Condorcet Fusion algorithm, achieved the best results. These results were achieved by combining three sensors: the text similarity sensor, the profile information sensor and the citation sensor. In this experiment, we are interested in determining the impact of the different sensors in out multisensor approach. To validate this, we separately tested our multisensor approach together with (i) the textual similarity and the profile sensors, (ii) the textual similarity and the citation sensors and (iii) the profile and the citation sensors. Table~\ref{tab:proximity-different-sensors} shows the obtained results. 

\begin{table}
\begin{center}
\resizebox{\columnwidth}{!} {
\begin{tabular}{ l c c c c c c}
									
~~										& ~~~~{\bf P@5}~~~~		& ~~~~{\bf P@10}~~~~	& ~~~~{\bf P@15}~~~~	& ~~~~{\bf P@20}~~~~	& ~~~~{\bf MAP}~~~~	\\
\hline
{\bf Sensors}																			&											&											&											&											&									\\
~~Text Similarity + Profile + Citation (T+P+C) 	& 0.7538							& 0.6385			  			& 0.5846							& 0.5615			 				& 0.4905	 		\\	
~~Text Similarity + Profile (T+P)					 				& 	0.7538		 				& 0.7000  						& 0.6564							& 0.6154							& 0.4961 						 \\
~~Text Similarity + Citation (T+C)			   	 		&  {\bf 0.8000}				& {\bf 0.7615}				& {\bf 0.7077}				& {\bf 0.6692}				& {\bf 0.5443}	 			 \\
~~Profile + Citation (P+C)								 			& 0.4769	 						& 0.4615							& 0.4359							& 0.4308							& 0.3828 						 \\
\hline
{\bf Relative Improvements} 											& 										& 										& 										& 										&			 				\\
~~T+C vs T+P+C										 								& 	+6.13\%  					& 	+19.26\%*					& 	+10.45\%*					&	+19.18\%*						& +10.97\%*				 \\
~~T+C vs T+P											   	 						& 	+6.13\%						&	+8.79\%							& 	+10.45\%					& 	+7.74\%* 					&	+9.72\%*	 	 \\
~~T+C vs P+C													 						& 	+67.75\%*					& 	+65.00\%*					&  +55.79\%*					& 	+55.34\%*					& 	+42.19\%*	 \\
\hline
\end{tabular}
}
\end{center}
\caption{The results obtained by making different combinations with the textual similarity sensor, the profile information sensor and the citation sensor in the proximity dataset.}
\label{tab:proximity-different-sensors}
\end{table}

Table~\ref{tab:proximity-different-sensors} shows that the best results were achieved when the text similarity sensor works together with the citation sensor. This means that the presence of the query topics in the author's document evidences together with information of the author's impact in the scientific community plays an important role to determine if some author is an expert in some specific topic. The results also show that taking into account the publication record of the authors does not contribute for the expert finding task in such framework. 

The significance tests performed show that the improvements achieved by the text similarity sensor together with the citation sensor are statistically more significant, in terms of MAP, than all the other combinations of sensors tested. Thus, the text sensor and the citation sensor acquired an improvement of more than 42\% over the profile sensor combined with the citation sensor, demonstrating their effectiveness.

\subsection{Performance of the MultiSensor Approach in the Enriched DBLP Dataset}~\label{subsec:3}

The previous results demonstrated the effectiveness of our multisensor approach using the Dempster-Shafer theory over poor datasets. In this experiment, we are concerned with the performance of our multisensor approach in enriched datasets, more specifically in the enriched version of the DBLP dataset. The results are illustrated in table~\ref{t4}.

\begin{table}
\resizebox{\columnwidth}{!} {
\begin{tabular}{ l c c c c c c }
												 										&~~~~{\bf P@5}~~~~ 		& ~~~~{\bf P@10}~~~~ 	& ~~~~{\bf P@15}~~~~ 	& ~~~~{\bf P@20}~~~~ 	& ~~~~{\bf MAP}~~~~ 	\\
\hline
{\bf MultiSensor Approach using}											&										&											&										&										&		\\
~~Dempster-Shafer + CombSUM (D-S + CSUM)					& {\bf 0.6462}			& {\bf 0.6000}				& {\bf 0.5590}			& {\bf 0.5385} 			& 0.3952		\\
~~Dempster-Shafer + Borda Fuse (D-S + BFuse)					& 0.2769						& 0.2615  						& 0.2564 						& 0.2500 						& 0.2713 \\
~~Dempster-Shafer + Condorcet Fusion (D-S + CFusion)	& 0.6308						& 0.5923 			 				& 0.5487	 					& 0.5269						& {\bf 0.4055}	\\
\hline
{\bf Standard Rank Aggregation Approach using} 				& 									& 										& 									& 									&			 \\
~~CombSUM																							& 0.3692						& 0.3308							& 0.3692						& 0.3308 						& 0.3073	\\
~~Borda Fuse																					& 0.2615						& 0.2692  						& 0.2667 						& 0.2692 						& 0.3169  \\
~~Condorcet Fusion																		& 0.4000						& 0.3538  						& 0.3436 						& 0.3154 						& 0.2773  \\
\hline
{\bf Relative Improvements}  													& 									& 										& 									& 									&			 	\\
~~ D-S + CFusion vs CombSUM 													&	+70.86\%					&	+79.05\%						&	+52.10\%					&	+59.08\%	 				& 	+31.96\%		\\
~~ D-S + CFusion vs Borda Fuse 												&	+141.23\%					&	+105.41\%						&	+120.02\%					&	+95.73\%					&	+27.96\%	 	\\
~~ D-S + CFusion vs Condorcet Fusion  								&	+57.70\%					&	+55.11\%						&	+67.41\%					&	+67.06\%					&	+46.23\%	  	\\
\hline
\end{tabular}
}
\caption{Results of the Dempster-Shafer theory of evidence combined with representative data fusion algorithms on an enriched version of the DBLP dataset.}
\label{t4} 
\end{table}

Table~\ref{t4} shows that the best performing approach was our multisensor approach together with the Condorcet Fusion algorithm. This approach achieved an improvement of more than 46\% when compared with the standard rank aggregation approach using the same fusion algorithm. Although, our best performing multisensor fusion approach using Dempster-Shafer outperformed all other standard rank aggregation methods, we cannot conclude that our approach is better, since there were no statistical significances detected.

In a separate experiment, we also tried to determine which combinations of sensors provided the best results for this dataset. Table~\ref{tab:dblp-impact-sensors} shows the obtained results.

\begin{table}
\resizebox{\columnwidth}{!} {
\begin{tabular}{ l c c c c c c}
																				 & ~~~~{\bf P@5}~~~~		& ~~~~{\bf P@10}~~~~	& ~~~~{\bf P@15}~~~~	& ~~~~{\bf P@20}~~~~	& ~~~~{\bf MAP}~~~~	\\
\hline
{\bf Sensors}															&											&											&											&											&									\\
~~Text Similarity + Profile + Citation~~  & 0.6308*						& 0.5923 			 				& 0.5487*							& 0.5269*							& 0.4055*					\\
~~Text Similarity + Profile							 		& 0.6615						& 0.6077				  		& {\bf 0.5846~}				& {\bf 0.5731}~				& {\bf 0.4530}~ 	\\
~~Text Similarity + Citation				 	 		& {\bf 0.6769}			& {\bf 0.6154}				& 0.5692~							& 0.5500~  						& 0.4157*	 				\\
~~Profile + Citation										 	& 0.5692 						& 0.5154* 						& 0.4462* 						& 0.4077* 						& 0.3454*					 \\
\hline
\end{tabular}
}
\caption{The results obtained by combining different sensors.}
\label{tab:dblp-impact-sensors}
\end{table}

In this experiment, the best results were achieved when the text similarity sensor is combined with the profile information sensor. This shows that, for this specific dataset, the presence of the query topics in the author's publication titles and abstracts together with the author's publication records, are very strong estimators of expertise. Thus, this information is vital to determine if someone is an expert in some topic. One can also see that the combination of the profile sensor with the citation sensor achieved the poorest results. These results are the same as the ones reported for the Proximity dataset, in Table~\ref{tab:proximity-different-sensors}. 

In this experiment, the improvements of the best performing sensors (text similarity and profile) were statistically more significant than the combination of all remaining sensors, in terms of MAP, this way showing the effectiveness of these estimators of expertise.

\subsection{Comparison with State of the Art}~\label{subsec:5}

In this experiment, we were concerned with the impact of our multisensor approach in the state of the art. Table~\ref{tab:comparison-with-state-of-the-art} presents the results of the baseline models proposed by~\citep{Balog06FormalModels}, namely the candidate-based Model 1 and the document-based Model 2. In order to make the comparison fair, we used the code made publicly available by K. Balog at \url{http://code.google.com/p/ears/}. Experiments revealed that Model 1 and Model 2 have a similar performance in such dataset, but achieved a lower performance when compared to the multisensor approach. In Model 1, when an author publishes a paper which contains a set of words which exactly match the query topics, this author achieves a very high score in this model. In addition, since we are dealing with very big datasets, there are lots of authors in such situation and consequently the top ranked authors are dominated by non-experts, while the real experts will be ranked lowered. In Model 2, since we only contain the publication's titles and some abstracts, the query topics might not occur very often in publications associated to expert authors. In such Model 2, the final ranking of a candidate is given by aggregating the scores that he achieved in each publication. If the document's abstract or title does not contain or is very poor in query topics, then the candidate will receive a lower score in the final ranking list. Our multisensor approach outperformed these state of the methods, because it enables the combination of various sources of evidence instead of just using textual similarities between query terms and documents.

\begin{table}
\begin{center}
\resizebox{\columnwidth}{!} {
\begin{tabular}{ l c c c c c c}

~~{\bf Approaches}	& ~~~~{\bf P@5}~~~~		& ~~~~{\bf P@10}~~~~	& ~~~~{\bf P@15}~~~~	& ~~~~{\bf P@20}~~~~	& ~~~~{\bf MAP}~~~~	\\
\hline
~~Model 1~\citep{Balog06FormalModels}	& 0.2769	& 0.2692					& 0.2616					&	0.2500				&	0.2715			 \\
~~Model 2~\citep{Balog06FormalModels}	& 0.2769	& 0.2846					& 0.2513					&	0.2449				&	0.2668		 		\\
\hline
\hline
~~{\bf MultiSensor Fusion using Dempster-Shafer theory}	&  {\bf 0.6615}		& {\bf 0.6077}			& {\bf 0.5846}				& {\bf 0.5731}					& {\bf 0.4530}	  \\  	
\hline

\end{tabular}
}
\end{center}
\caption{ Comparison of our best MultiSensor approach (Profile + Text Sensors) with other state of the art approaches which use the computer science DBLP dataset}
\label{tab:comparison-with-state-of-the-art}
\end{table}

\subsection{Comparison with State of the Art Supervised Approaches}~\label{subsec:6}

In the task of expert finding, there have been several effective approaches proposed in the literature, exploring different retrieval models and different sources of evidence for estimating expertise. More recently, some works have been proposed in the literature which use supervised machine learning techniques to combine different sources of evidence in an optimal way~\citep{Yang09bole,Macdonald11aggr}. In this section, we reproduce the experiments of some of those works that use Learning to Rank algorithms to effectively combine different estimators of expertise. More specifically, we applied the SVMmap and SVMrank to our test set in order to determine the impact of our multisensor approach against a supervised one.

The general idea of Learning to Rank is to use hand-labelled data to train ranking models, this way leveraging on data to combine the different estimators of relevance in  an optimal way. In the training process, the learning algorithm attempts to learn a ranking function capable of sorting experts in a way that optimizes a bound of an information retrieval performance measure (e.g. Mean Average Precision), or which tries to minimize the number of misclassifications between expert pairs, or which even tries to directly predict the relevance scores of the experts. In the test phase, the learned ranking function is applied to determine the relevance between each expert towards a new query.

The two algorithms tested were SVMmap and SVMrank. The SVMmap method~\citep{Yue07Support} builds a ranking model through the formalism of structured Support Vector Machines~\citep{Tsochantaridis05Structured}, attempting to optimize the metric of Average Precision (AP). The basic idea of SVMmap is to minimize a loss function which measures the difference between the performance of a perfect ranking (i.e., when the Average Precision equals one) and the minimum performance of an incorrect ranking. 

The SVMrank method~\citep{Joachims06rank} also builds a ranking model through the formalism of Support Vector Machines. However, the basic idea of SVMrank is to attempt to minimize the number of misclassified expert pairs in a pairwise setting. This is achieved by modifying the default support vector machine optimization problem, by constraining the optimization problem to perform the minimization of the number of misclassified pairs of experts. 

Since the proximity dataset contains very sparse data, due to its lack of information, the task of expert finding in this dataset would be very trivial using a supervised machine learning approach. Since many non relevant authors don't have many information associated to them, it would be easy to find a hyperplane which would be able to classify an author as being relevant or non relevant. For this reason, we performed a supervised machine learning test in the enriched dataset in order to make the task more difficult. Table~\ref{tab:comparison-with-supervised-dblp} presents the results of these two algorithms, for the enriched DBLP dataset, and their comparison against our multisensor Fusion approach using the Dempster-Shafer theory of evidence.

\begin{table}
\begin{center}
\resizebox{\columnwidth}{!} {
\begin{tabular}{ l c c c c c c}

{\bf Approaches}									& {\bf P$@$5}		& {\bf P$@$10}	& {\bf P$@$15}	& {\bf P$@$20} 	& {\bf MAP}  \\
\hline
{\bf MultiSensor Fusion using Dempster-Shafer}		& {\bf 0.6308}		& {\bf 0.5923} 	& {\bf 0.5487}	& {\bf 0.5269}	& 0.4055	\\	  
~~SVMmap~\citep{moreira11epia}						& 0.4292			&	0.4313		&	0.4014		&	0.4042		& 0.4068	\\	
~~SVMrank~\citep{moreira11epia}						& 0.4583			& 0.4750		&	0.4958		& 0.4642	& {\bf 0.4289}	\\	
\hline
MultiSensor Approach vs SVMmap						& +46.97\%			& +37.33\%		& +36.70\%		& +30.36\%		& -0.31\%	\\
MultiSensor Approach vs SVMrank						& +37.64\%			& +24.69\%		& +10.67\%		& +13.51\%		& -5.46\%	\\
\hline
\end{tabular}
}
\end{center}
\caption{ Comparison of our multisensor approach using the three sensors against two supervised methods of the state of the art for the dblp dataset}
\label{tab:comparison-with-supervised-dblp}
\end{table}

Table~\ref{tab:comparison-with-supervised-dblp} shows that the results obtained in the different algorithms revealed slightly variations when concerning the Mean Average Precision Metric, leading to the conclusion that the application of machine learning techniques to this dataset do not bring great advantages. In addition, our multisensor approach achieved better results than the supervised learning to rank algorithms for the $P@k$ performance measure. This metric is very important in the task of expert finding in digital libraries, since the user is only interested in searching for the top $k$ relevant experts of some topic. 

\begin{figure}
\resizebox{\columnwidth}{!} {
\includegraphics{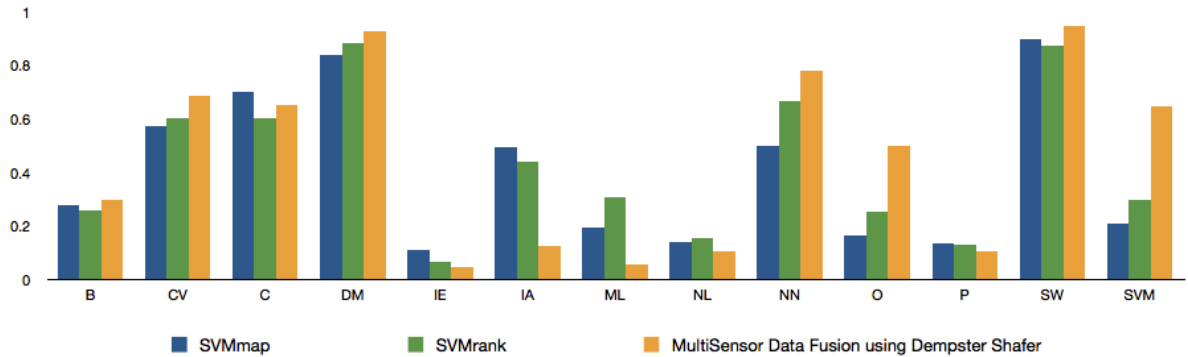}
}
\caption{Average precision over the different query topics for the supervised learning to rank algorithms  SVMmap and SVMrank and for our sensor fusion approach using the Dempster-Shafer theory for the enriched DBLP dataset}
\label{fig:dblp-vs-supervised}
\end{figure}

Since the statistical significance tests performed did not accuse any differences between the algorithms, then we can state that our multisensor approach achieves a performance similar to supervised machine learning techniques and it even has the advantage of not requiring hand-labelled data with personal relevance judgements. This means that the proposed method is general and can be scalable to a real world scenario, while machine learning approaches can only be trained with a small set of data which is not representative of a real expert finding scenario. Finally, Figure~\ref{fig:dblp-vs-supervised} supports the above observations, showing the Average Precision of the three algorithms for each query. One can easily see that the results obtained show slightly variations between the algorithms demonstrating that our method achieves a similar performance to the supervised learning to rank algorithms.

\section{Conclusion}\label{sec:conclusions}

We proposed a multisensor Data Fusion approach using the Dempster-Shafer theory of Evidence together with Shannon's Entropy. In order to extract different sources of expertise from these documents, we defined three sensors: a text similarity sensor, a profile information sensor and a citation sensor. The text sensor collects events derived from traditional information retrieval techniques, which measure term co-occurrences between the query topics and the documents associated to a candidate. The profile sensor measures the total publication record of a candidate throughout his career, under the assumption that highly prolific candidates are more likely to be considered experts. And the citation sensor uses citation graphs to capture the authority of candidates from the attention that others give to their work in the scientific community. 

Experimental results revealed that the Dempster-Shafer theory of evidence combined with the Shannon's entropy formula is able to address an important issue which so far has been ignored in the expert finding literature: conflicting information. When merging information from different sources of evidence, there will always be significant levels of uncertainty. These uncertainty levels arise because we only have a partial knowledge of the state of the world. The Dempster-Shafer framework can lead with these kind of problems through the usage of a combination rule which measures the total amount of conflicts between these three sensors. This way a final accurate decision can be made from noisy data.

We compared our multisensor approach against representative approaches of the state of the art of expert finding. Our method showed great improvements, demonstrating that the levels of uncertainty provide a very important issue, not only for expert finding, but also for general ranking problems of information retrieval. It was interesting to notice that our approach has also a good performance on datasets which lack on information, showing that the Dempster-Shafer theory of evidence can be also effective in poor datasets, making the approach scalable to any information retrieval tasks such as entity ranking or in search engines in order to address the problem of uncertainty.

Finally, we tested our algorithm by comparing it to supervised machine learning approaches of the state of the art which use supervised learning to rank techniques. Although these approaches usually provide good results for this task, they suffer from the disadvantage of the lack of hand-labelled data with relevance judgements about the level of expertise of an author towards a query. Even the term expert is hard to define, since the expertise areas of a candidate are difficult to quantify and the experience of a candidate is always varying through time. As a consequence,  this hand-labelled data will only be the evidence of the trainers judgements and the final system will only reflect their biases. In this scope, our multisensor approach is more general and more useful than supervised approaches, since it does not require the training of a system and therefore it is much faster to implement and scalable to a real world scenario. Our method is more focused in finding information from the data given, rather than finding patterns based on personal relevance judgements and this is a major advantage towards supervised learning to rank approaches.







\end{document}